\pdfoutput=1
\documentclass[11pt]{article}

\usepackage{acl}

\usepackage{times}
\usepackage{latexsym}

\usepackage[T1]{fontenc}
\usepackage[utf8]{inputenc}

\usepackage{microtype}

\usepackage{inconsolata}

\usepackage{graphicx}
\usepackage{inconsolata}
\usepackage{microtype}
\usepackage{multirow}
\usepackage{multicol}
\usepackage{makecell}
\usepackage[skins]{tcolorbox}
\usepackage{amsmath}
\usepackage{amsfonts}
\usepackage{adjustbox}
\usepackage{bm}
\usepackage{enumitem}
\usepackage{ulem}
\usepackage{booktabs}
\usepackage{soul}
\usepackage{bbding}
\usepackage{pifont}
\usepackage[percent]{overpic}
\usepackage{colortbl}

\definecolor{pastelblue}{HTML}{A1C9F4}
\definecolor{pastelorange}{HTML}{FFB482}
\definecolor{pastelgreen}{HTML}{8DE5A1}
\definecolor{pastelred}{HTML}{FF9F9B}
\definecolor{pastelpurple}{HTML}{D0BBFF}
\definecolor{pastelgold}{HTML}{FFD700}
\definecolor{pastelpink}{HTML}{FF69B4}

\colorlet{pastelbluemuted}{pastelblue!75}
\colorlet{pastelorangemuted}{pastelorange!75}
\colorlet{pastelgreenmuted}{pastelgreen!75}
\colorlet{pastelredmuted}{pastelred!75}
\colorlet{pastelpurplemuted}{pastelpurple!75}
\colorlet{pastelgoldmuted}{pastelgold!75}
\colorlet{pastelpinkmuted}{pastelpink!75}

\def \fmargin {-10pt}

\definecolor{red}{RGB}{255, 0, 0}
\definecolor{green}{RGB}{0, 176, 80}
\definecolor{purple}{RGB}{128, 0, 128}
\newcommand{\cmark}{\textcolor{green}{\ding{51}}} % ✔
\newcommand{\xmark}{\textcolor{red}{\ding{55}}}

\newcommand{\customfootnotetext}[2]{{
  \renewcommand{\thefootnote}{#1}
  \footnotetext[0]{#2}}}

\title{\textsc{Radar}: Enhancing Radiology Report Generation with Supplementary Knowledge Injection}
\author{Wenjun Hou$^{1,2}$, Yi Cheng$^{1\ast}$, Kaishuai Xu$^{1\ast}$, Heng Li$^{2}$, Yan Hu$^{2\dagger}$, Wenjie Li$^{1}$, Jiang Liu$^{2,3\dagger}$ \\
$^1$Department of Computing, The Hong Kong Polytechnic University \\
$^2$Research Institute of Trustworthy Autonomous Systems and \\Department of Computer Science and Engineering, \\
Southern University of Science and Technology \\
$^3$School of Computer Science, University of Nottingham Ningbo China \\
\texttt{houwenjun060@gmail.com}
}

\begin{document}
\maketitle
\customfootnotetext{$\ast$}{Equal contribution.}
\customfootnotetext{$\dagger$}{Corresponding authors.}
\begin{abstract}
Large language models (LLMs) have demonstrated remarkable capabilities in various domains, including radiology report generation. Previous approaches have attempted to utilize multimodal LLMs for this task, enhancing their performance through the integration of domain-specific knowledge retrieval. However, these approaches often overlook the knowledge already embedded within the LLMs, leading to redundant information integration. To address this limitation, we propose \textsc{Radar}, a framework for enhancing radiology report generation with supplementary knowledge injection. \textsc{Radar} improves report generation by systematically leveraging both the internal knowledge of an LLM and externally retrieved information. Specifically, it first extracts the model's acquired knowledge that aligns with expert image-based classification outputs. It then retrieves relevant supplementary knowledge to further enrich this information. Finally, by aggregating both sources, \textsc{Radar} generates more accurate and informative radiology reports. Extensive experiments on MIMIC-CXR, \textsc{CheXpert-Plus}, and \textsc{IU X-ray} demonstrate that our model outperforms state-of-the-art LLMs in both language quality and clinical accuracy\footnote{Our code is available at: \url{https://github.com/wjhou/Radar}}. 
\end{abstract}

\begin{figure}[t]
    \centering
    \setlength\belowcaptionskip{\fmargin}
    \includegraphics[width=1.0\linewidth]{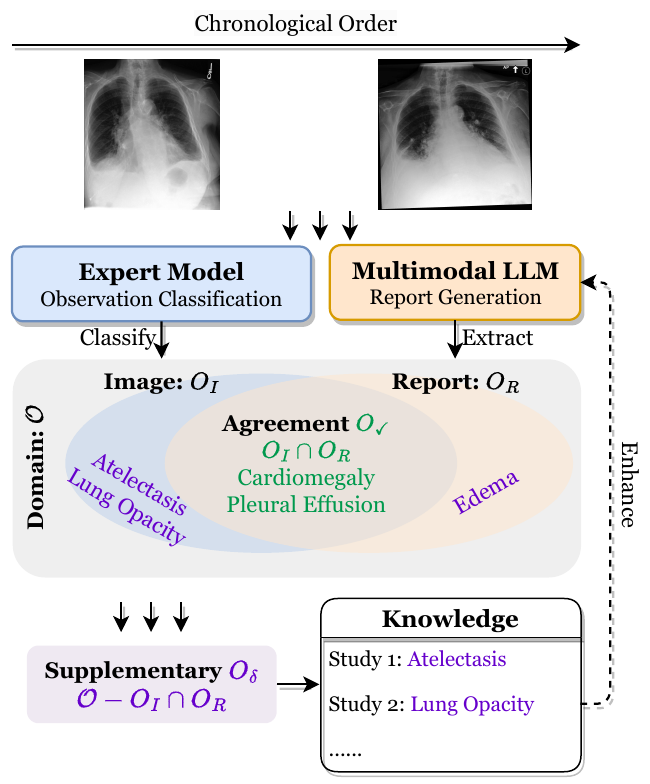}
    \caption{A motivating example. The report directly generated by the multimodal LLM showcases its knowledge regarding several findings ($O_R$) but can contain hallucinations and overlook some other findings. To address this, we regard the part that aligns with another expert model ($O_R \cap O_I$) as trustworthy and we incorporate supplementary knowledge for the remaining part ($\mathcal{O}-O_R \cap O_I$) to enhance the report generation.}
    \label{figure: figure1}
\end{figure}

\section{Introduction}
Radiology report generation \cite{r2gen,r2gencmn} plays a crucial role in chest X-ray interpretation, requiring highly specialized domain knowledge \cite{radgraph,chexpert}. Recent advances in foundation models \cite{radialog,chexagent,maira1}, which leverage large language models (LLMs) for enhanced medical image analysis, have demonstrated remarkable potential in generating fluent and cohesive clinical text, aiding radiologists in their diagnostic workflow.

Despite their ability to generate highly readable and clinically plausible report content, LLMs still face persistent challenges in ensuring clinical accuracy. One major challenge lies in the knowledge gap between the medical and general domains. Many studies have attempted to bridge this disparity by augmenting models with retrieved domain-specific knowledge \cite{mia,ppked,dcl,ranjit2023retrievalaugmentedchestxray,sun-etal-2025-fact}. However, these approaches often overlook the knowledge LLMs have already acquired. That is, much of the retrieved information is often duplicate knowledge already encoded within the model's parameters, leading to redundant information retrieval. Moreover, the knowledge learned by LLMs \cite{r2llm} is not always trustworthy, as hallucinations frequently occur \cite{hallucination}. For instance, in Figure \ref{figure: figure1}, the LLM correctly identifies \textit{Cardiomegaly}, making the retrieval of additional knowledge about this observation unnecessary. Additionally, the generated \textit{Pleural Effusion} is highly credible, as it aligns with the expert model, whereas \textit{Edema} remains uncertain. Thus, balancing learned and retrieved knowledge in radiology report generation is crucial to address these challenges.

In this paper, we propose \textsc{Radar}, a framework for \underline{\textsc{Rad}}iology report generation that integrates both the internal knowledge of LLMs and external supplement\underline{\textsc{ar}}y knowledge. Our framework primarily consists of two stages: preliminary findings generation and supplementary findings augmentation. In the first stage, \textsc{Radar} generates an initial report from the input images. Subsequently, an expert model processes the images for observation classification. The overlapping information between the generated report and the classified observations is identified as high-confidence internal knowledge. In the second stage, \textsc{Radar} additionally retrieves new knowledge to supplement the internal knowledge. Finally, both internal and supplementary knowledge sources are aggregated to enhance the report generation process. Our main contributions can be summarized as follows:
\begin{itemize}
    \item We propose \textsc{Radar}, a novel framework that enhances the clinical accuracy of radiology report generation by effectively integrating both the internal knowledge of LLMs and externally retrieved domain-specific knowledge.
    \item To optimize knowledge utilization, we introduce a knowledge extraction method that identifies and retains non-overlapping information from the model’s learned knowledge, reducing redundancy and bridging the knowledge gap.
    \item We conduct extensive experiments on three benchmark datasets: MIMIC-CXR, \textsc{CheXpert-Plus}, and \textsc{IU X-ray}, demonstrating the effectiveness of \textsc{Radar}.
\end{itemize}
\begin{figure*}[t]
    \centering
    \includegraphics[width=1.0\linewidth]{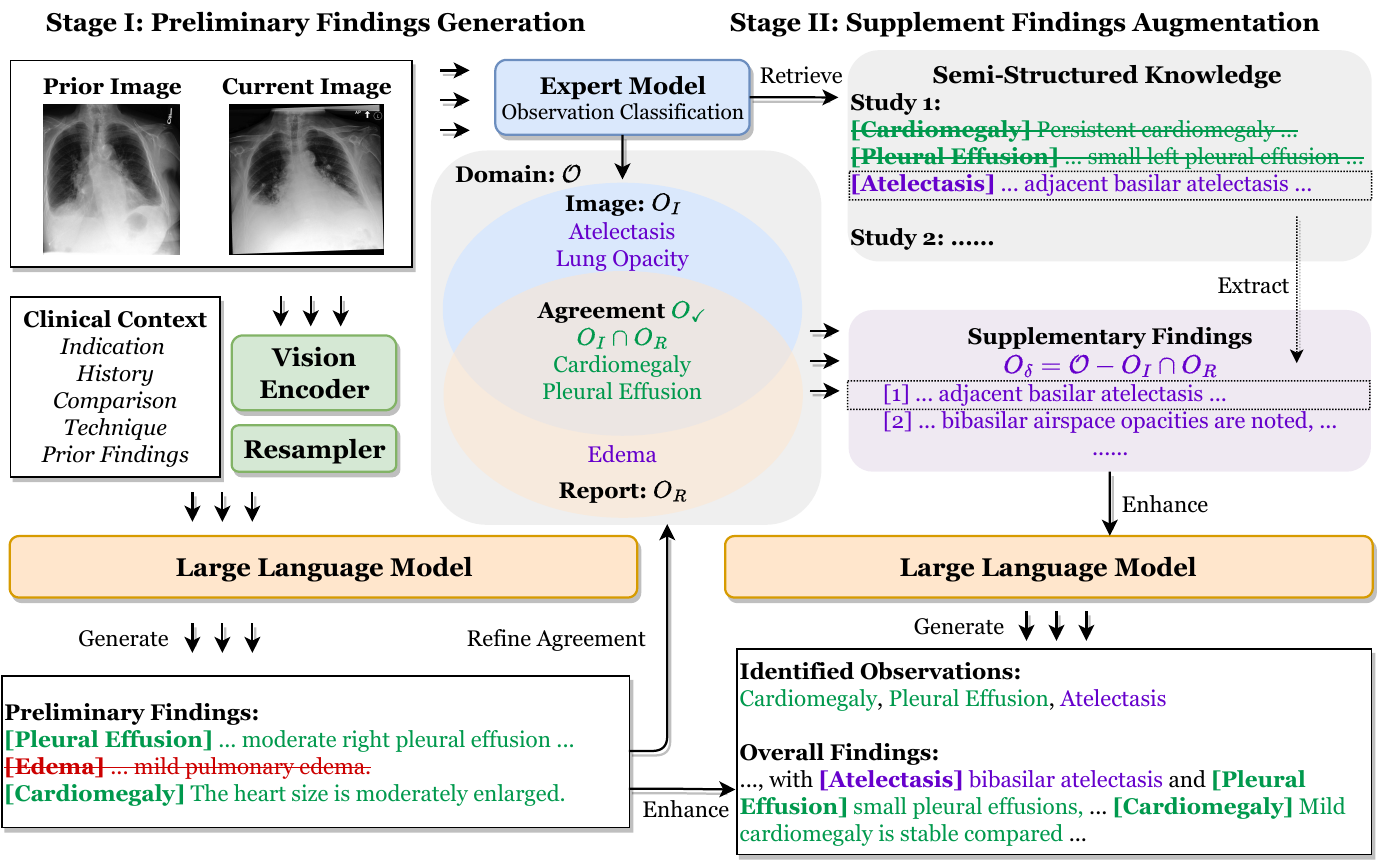}
    \caption{Overview of the \textsc{Radar}. In Preliminary Findings, only sentences that reach agreement are retained, whereas in Supplementary Findings, only sentences that supplement the Preliminary Findings are preserved.}
    \label{figure2}
\end{figure*}
\section{Preliminary}
\subsection{Problem Formulation}
A multimodal LLM (MLLM) generally consists of a vision encoder, a vision connector that transforms visual signals into the language space (e.g., MLP \cite{llava}, Q-Former \cite{blip2}, or Perceiver Resampler \cite{blip3}), and an LLM, as illustrated in the left part of Figure \ref{figure2}. For radiology report generation\footnote{In this paper, "report" typically refers to "findings," and we use these two terms interchangeably.}, the MLLM takes a radiograph $X$, its prior $X_p$ (if available), and the clinical context $C$ (e.g., \textit{Indication} or \textit{Prior Findings}) as input and generates the report $Y = \{y_1, \dots, y_T\}$. The probability of the $t$-th token is computed as follows:
\begin{equation*}
    p(y_t) = \mathrm{MLLM}(X, X_p, C, y_{<t}),
\end{equation*}
where the MLLM is optimized using the negative log-likelihood loss:
\begin{equation*}
    \mathcal{L} = -\sum_{t=1}^{T} \log p(y_t).
\end{equation*}

\subsection{Semi-Structured Report as Knowledge}\label{semi-know}
In this paper, the training set of \textsc{MIMIC-CXR} serves as the knowledge source for radiology report generation. To effectively leverage the knowledge encoded in each report, we convert it into semi-structured data. Specifically, given a report consisting of $N$ sentences, $Y = \{S_1, \dots, S_N\}$, we annotate each sentence using the 14-category CheXpert observations \cite{chexpert} with the CheXbert model \cite{chexbert}. Each observation falls into one of four classes: \textit{Positive}, \textit{Negative}, \textit{Uncertain}, or \textit{Blank}. To ensure conciseness, we retain only sentences annotated with \textit{Positive} observations. These selected sentences collectively represent the knowledge extracted from the report, as illustrated in the top-right part of Figure \ref{figure2}. Note that we annotate and process Preliminary Findings (\S \ref{section: PF}) and Supplementary Findings (\S \ref{section: SF}) in the same manner.

\section{The \textsc{Radar} Framework}
\subsection{Stage I: Preliminary Findings Generation}\label{section: PF}
We illustrate the Stage I process in the left part of Figure \ref{figure2}. To assess the learned knowledge of an LLM, we first feed the input ($X$, $X_p$, and $C$) into \textsc{Radar} to generate a report $\hat{Y}$:
\begin{equation*}
    \hat{Y} =\underset{{\hat{Y} \in \mathcal{Y}}}{\mathrm{argmax}}\; \prod^{T}_{t=1}\mathrm{MLLM}(X, X_p, C, \hat{y}_{<t}),
\end{equation*}
where $\mathcal{Y}$ represents the set of possible reports. Note that exact maximization is intractable and we employ an approximate decoding algorithm for generation. Next, we convert the findings into semi-structured knowledge, as described in \S \ref{semi-know}, and denote the observations of $\hat{Y}$ as $O_R$. 

To extract credible knowledge from $\hat{Y}$ while filtering out untrustworthy information, we train an expert model that predicts observations for the image. Unlike previous works \cite{organ,radialog}, which consider only the image as input, we incorporate the clinical context to enhance performance. Specifically, the expert model $f(X)$ encodes $X$ and $C$ using an image encoder $\mathrm{Encoder}_v$ and a text encoder $\mathrm{Encoder}_t$, respectively, and then processes their outputs through an MLP for observation classification:
\begin{equation*}
    \mathbf{h}_v = \mathrm{Encoder}_v(X),\quad\mathbf{h}_t = \mathrm{Encoder}_t(C),
\end{equation*}
\begin{equation*}
    p(O_i) = \sigma(\mathrm{MLP}([\mathbf{h}_v; \mathbf{h}_t])),
\end{equation*}
where $[;]$ is the concatenation function, $\mathbf{h}_v$ and $\mathbf{h}_t$ are the pooled outputs of the image and text encoders, respectively, and $p(O_i)$ represents the probability of the $i$-th observation. We denote the observations derived from $f(X)$ as $O_I$, and the credible and high-confidence observations, $O_\checkmark$, are then obtained by intersecting $O_I$ and $O_R$, as follows:
\begin{equation*}
    O_\checkmark = O_I \cap O_R.
\end{equation*}
Finally, we refine $\hat{Y}$ by removing sentences that do not correspond to $O_\checkmark$, yielding the {Preliminary Findings (PF)}. 

To train the expert model, we collect observations from each report as image annotations and optimize the expert model using binary cross-entropy loss. Following \citet{radialog}, we address data imbalance by re-weighting the positive observations with a log-scale weight, defined as $\alpha_{i} = \log \left( 1 + \frac{|\mathcal{D}_{\text{train}}|}{w_i} \right)$, where $|\mathcal{D}_{\text{train}}|$ is the total number of training samples and $w_i$ denotes the frequency of observation $O_i$.

\subsection{Stage II: Supplementary Findings Augmentation}\label{section: SF}
\noindent\textbf{Supplementary Knowledge Retrieval.}  
We follow the retrieval process of \citet{mia} to search for domain knowledge. Specifically, the expert model described in \S \ref{section: PF} produces probabilities for 14 observations, and we compute the similarity between different samples using KL-divergence:
\begin{equation*}
    \hat{z} = \mathrm{Normalize}(f(X)),
\end{equation*}
\begin{equation*}
    \mathrm{Sim}(X, X_i) = -\sum_{j=1}^{|\mathcal{O}|} \hat{z}_j \log \frac{\hat{z}_j}{\hat{z}_{i,j}},
\end{equation*}
where $\mathrm{Normalize(\cdot)}$ normalizes $f(X)$ to $1$, $\hat{z}$ represents the normalized scores for $(X)$, and $\hat{z}_{i,j}$ denotes the score of the $j$-th observation in the $i$-th sample from the database (i.e., the training set of the MIMIC-CXR dataset). We then rank the samples based on their similarity scores, $\mathrm{Sim}(X, X_i)$, and retrieve the top-$K$ reports, denoted as $\mathcal{Y}^{S} = \{Y^S_1, \dots, Y^S_K\}$.

\noindent\textbf{Supplementary Knowledge Extraction.}  
Since the retrieved information may overlap with the knowledge learned by LLMs, we extract only supplementary knowledge based on two principles:  
(1) it should be concise and relevant, and  
(2) it should complement, rather than duplicate, the preliminary findings. Thus, for each supplementary report $Y_i^S$ with its corresponding observations $O^S$, we retain only the following observations:
\begin{equation*}
    O_\delta = \mathcal{O} - O_\checkmark.
\end{equation*}
Next, we convert $Y^{S}_i$ into semi-structured knowledge and remove sentences that do not correspond to $O_\delta$, referring to these findings as {Supplementary Findings (SF)}. Notably, all sentences corresponding to negative observations are removed, ensuring that SF remains concise and clinically relevant.

\subsection{Enhanced Radiology Report Generation}\label{section: OI}
We integrate both PF and SF into the clinical context $C$ to form the augmented context $C^A$, from which the final report $Y$ is generated as:
\begin{equation*}
    {Y} =\underset{{{Y} \in \mathcal{Y}}}{\mathrm{argmax}}\; \prod^{T}_{t=1}\mathrm{MLLM}(X, X_p, C^A, {y}_{<t}).
\end{equation*}
Since PF and SF contain information from various studies, summarizing high-level information before generating the report is necessary. Thus, we include the observations of $Y$ as part of the training targets. Specifically, during training, $Y$ is converted into a structured format:
\begin{equation*}
    Y^O = \{O_1, \dots, O_N, y_1, \dots, y_L\},
\end{equation*}
where $\{O_1, \dots, O_N\}$ represents the observations in $Y$, and $\{y_1, \dots, y_L\}$ corresponds to the tokens of the report. We refer to this process as {Observation Identification (OI)}. During inference, we extract the final report from the generated output for evaluation.
\begin{table*}[t]
    \centering
    \resizebox{\textwidth}{!}{
    \begin{tabular}{l|cccc|ccccccc}
    \toprule
    \multicolumn{12}{c}{\textbf{Dataset: MIMIC-CXR} (\textit{Training and Evaluation})} \\
    \midrule
    \multirow{2}{*}{\textbf{Model}} & \multicolumn{4}{c|}{\textbf{Lexical Metrics}} & \multicolumn{7}{c}{\textbf{Clinical Metrics} (\textit{CheXpert: Uncertain as Negative / \textcolor{gray}{\textit{Positive}}})} \\
    & \textbf{B-1} & \textbf{B-4} & \textbf{MTR} & \textbf{R-L} & \textbf{RG-F}$_1$ & \textbf{RG}$_\text{ER}$ & \textbf{CliQ}$_\text{0}$ ($\downarrow$) & \textbf{$^{14}$Ma-F}$_1$ & \textbf{$^{5}$Ma-F}$_1$ & \textbf{$^{14}$Mi-F}$_1$ & \textbf{$^{5}$Mi-F}$_1$ \\ 
    \midrule
    RadFM & $-$ & $0.128$ & $-$ & $0.182$ & $-$ & $-$ & $-$ & $-$ & $-$ & $-$ & $-$\\
    XrayGPT & $0.128$ & $0.004$ & $0.079$ & $0.111$ & $-$ & $-$ & $-$ & $-$ & $-$ & $-$ & $-$ \\
    R2GenGPT & $0.411$ & $0.134$ & $0.160$ & $0.297$ & $-$ & $-$ & $-$ & $0.389$ & $-$ & $-$ & $-$\\
    R2-LLM & $0.402$ & $0.128$ & $0.175$ & $0.291$ & $-$ & $-$ & $-$ & $-$ & $-$ & $-$ & $-$ \\
    RaDialog & $0.346$ & $0.095$ & $0.140$ & $0.271$ & $-$ & $-$ & $-$ & $0.394$ & $-$ & $-$ & $-$ \\
    LLaVA-Med & $0.354$ & $0.149$ & $0.353$ & $0.276$ & $0.191$ & $0.238$ & $3.30$ & $0.269$ & $0.363$ & $0.427$ & $0.439$\\
    CheXagent & $0.169$ & $0.047$ & $-$ & $0.215$ & $-$ & $0.205$ & $-$ & $0.247$ & $0.345$ & $0.393$ & $0.412$\\
    GPT-4V & $0.164$ & $0.178$ & $-$ & $0.132$ & $-$ & $0.132$ & $-$ & $0.204$ & $0.196$ & $0.355$ & $0.258$ \\
    Med-PaLM & $0.323$ & $0.115$ & $-$ & $0.275$ & $0.267$ & $-$ & $-$ & $0.398$ & $0.516$ & $0.536$ & $0.579$ \\
    LLaVA-Rad & $0.381$ & $0.154$ & $-$ & $0.306$ & $-$ & $0.294$ & $-$ & $0.395$ & $0.477$ & $0.573$ & $0.574$ \\
    \midrule
    MAIRA-1 & $0.392$ & $0.142$ & $0.333$ & $0.289$ & $0.243$ & $0.296$ & $3.10$ & \makecell{$0.386$\\\textcolor{gray}{$0.423$}} & \makecell{$0.477$\\\textcolor{gray}{$0.517$}} & \makecell{$0.557$\\\textcolor{gray}{$0.553$}} & \makecell{$0.560$\\\textcolor{gray}{$0.588$}}\\
    \midrule
    MAIRA-2 & $0.465$ & $0.234$ & $0.420$ & $\underline{0.384}$ &  $\bm{0.346}$ &  $\bm{0.396}$ & $\underline{2.64}$ & $\underline{0.416}$ & $0.504$ & $0.581$ & $0.591$ \\
    MedVerse & $-$ & $0.178$ & $-$ & $-$ & $0.280$ & $-$ & ${2.71}$ & $-$ & $-$ & $-$ & $-$ \\
    M4CXR & $0.339$ & $0.103$ & $-$ & $-$  & $0.218$ & $0.285$ & $-$ & $0.400$ & $0.495$ & $\underline{0.606}$ & $\underline{0.618}$ \\
    Libra &  $\bf{0.513}$ & $\underline{0.245}$ &  $\bf{0.489}$ & ${0.367}$ & ${0.329}$ & $0.376$ & ${2.70}$ & $0.404$ & $\underline{0.538}$ & $0.559$ & $0.601$\\
    \midrule
    \textsc{Radar} (Ours) & $\underline{0.509}$ & $\bf{0.262}$ & $0.450$ & $\bf{0.397}$ & $\bm{0.346}$ & $\underline{0.393}$ & $\bm{2.61}$ & 
    \makecell{$\bf{0.460}$\\\textcolor{gray}{$0.497$}} & \makecell{$\bf{0.567}$\\\textcolor{gray}{$0.602$}} & \makecell{$\bf{0.627}$\\\textcolor{gray}{$0.627$}} & \makecell{$\bf{0.653}$\\\textcolor{gray}{$0.674$}} \\
    \bottomrule
    \end{tabular}}
    \caption{Evaluation results of our model and baseline methods on the \textsc{MIMIC-CXR} dataset. Baseline results are cited from their respective literature. The best results are shown in {\textbf{bold}}, while {\underline{underlined}} values indicate the second-best results. $\downarrow$ denotes that lower values are better. \textcolor{gray}{Results of CheXpert} treat \textit{Uncertain} labels as \textit{Positive} when compared with MAIRA-1. Comparisons with SOTA specialists are provided in Table \ref{table: full_specialist_results}.}
    \label{table: mllm_mimic_frontal_results}
\end{table*}

\begin{table}[ht]
    \centering
    \resizebox{\linewidth}{!}
    {
    \begin{tabular}{l|cc|cccc}
    \toprule
    \multicolumn{7}{c}{\textbf{Dataset: \textsc{IU X-ray}} (\textit{Evaluation Only})} \\\midrule
    \multirow{2}{*}{\textbf{Model}} & \multicolumn{2}{c|}{\textbf{Lexical}} & \multicolumn{4}{c}{\textbf{Clinical}} \\
    & \textbf{B-4} & \textbf{R-L} & \textbf{RG-F}$_1$ & \textbf{CliQ$_\text{0}$} ($\downarrow$) & \textbf{$^{14}$Ma-F$_1$} & \textbf{$^{14}$Mi-F$_1$} \\ 
    \midrule
    LLaVA-Rad & $-$ & $0.253$ & $-$ & $-$ & $-$ & $0.535$ \\
    MAIRA-2 & ${0.117}$ & ${0.274}$ & $0.271$ & $2.68$ & $0.319$ & $0.525$ \\
    \midrule
    \textsc{Radar} (Ours) &  ${0.116}$ &  ${0.276}$ & $0.237$ & $2.78$ & ${0.325}$ & ${0.546}$ \\
    \textsc{Backbone} & $0.112$ & $0.275$ & $0.236$ & $2.79$ & $0.269$ & $0.514$ \\
    \bottomrule
    \end{tabular}}
    \caption{Evaluation on the \textsc{IU X-ray} dataset. Results of LLaVA-Rad and MAIRA-2 are cited from \citet{maira2}.}
    \label{table: iu_xray_results}
\end{table}

\begin{table}[ht]
    \centering
    \resizebox{\linewidth}{!}{
    \begin{tabular}{l|l|cc|ccc}
    \toprule
    \multicolumn{7}{c}{\textbf{Dataset: \textsc{CheXpert Plus}} (\textit{Evaluation Only})} \\\midrule
    \multirow{2}{*}{\textbf{Model}} & \multirow{2}{*}{\textbf{Train}} & \multicolumn{2}{c|}{\textbf{Lexical}} & \multicolumn{3}{c}{\textbf{Clinical}} \\ 
     & & \textbf{B-4} & \textbf{R-L} & \textbf{RG}$_{\overline{\text{ER}}(\text{ER})}$ & \textbf{$^{14(5)}$Ma-F$_1$} & \textbf{$^{14(5)}$Mi-F$_1$} \\ 
    \midrule
    \multirow{3}{*}{\textsc{Swin$_\text{v2}$-BERT}} & M$^\star$ & ${0.034}$ & ${0.191}$ & ${0.136}$ (${0.198}$) & ${0.268}$ (${0.383}$) & ${0.410}$ (${0.423}$) \\
    & C & $0.057$ & $0.228$ & $0.183$ ($0.250$) & $0.331$ ($0.401$) & $0.508$ ($0.432$) \\
    & M\&C & $0.056$ & $0.234$ & $0.201$ ($0.277$) & $0.366$ ($0.495$) & $0.560$ ($0.532$) \\
    \midrule
    \textsc{Radar} (Ours) & M & ${0.076}$ & ${0.203}$ & ${0.143}$ (${0.216}$) & \makecell{${0.362}$ (${0.417}$)\\\textcolor{gray}{$0.401$ ($0.540$)}} & \makecell{${0.541}$ (${0.524}$)\\\textcolor{gray}{$0.554$ ($0.608$)}}  \\
    \textsc{Backbone} & M & $0.073$ & $0.203$ & $0.143$ ($0.206$) & \makecell{$0.282$ ($0.437$) \\ \textcolor{gray}{$0.317$ ($0.502$)}} & \makecell{$0.477$ ($0.466$) \\ \textcolor{gray}{$0.492$ ($0.552$)}}\\
    \bottomrule
    \end{tabular}}
    \caption{Evaluation on the \textsc{CheXpert Plus} dataset. The results for \textsc{Swin$_\text{v2}$-BERT} are cited from \citet{chexpert_plus}, and we primarily compare \textsc{Radar} with its $\star$ variant. The "Train" column indicates the training datasets, where M and C denote the MIMIC-CXR and \textsc{CheXpert Plus} datasets, respectively.}
    \label{table: chexpert_plus_results}
\end{table}
\begin{table*}[t]
    \centering
    \resizebox{\textwidth}{!}{
    \begin{tabular}{l|ccc|cccc|ccccccc}
    \toprule
    \multirow{2}{*}{\textbf{Model}} & \multicolumn{3}{c|}{\textbf{Modules}} & \multicolumn{4}{c|}{\textbf{Lexical Metrics}} & \multicolumn{7}{c}{\textbf{Clinical Metrics} (\textit{CheXpert: Uncertain as Negative})} \\
    & \textbf{PF} & \textbf{SF} & \textbf{OI} & \textbf{B-1} & \textbf{B-4} & \textbf{MTR} & \textbf{R-L} & \textbf{RG-F}$_1$ & \textbf{RG}$_\text{ER}$ & \textbf{CliQ}$_\text{0}$ ($\downarrow$) & \textbf{$^{14}$Ma-F}$_1$ & \textbf{$^{5}$Ma-F}$_1$ & \textbf{$^{14}$Mi-F}$_1$ & \textbf{$^{5}$Mi-F}$_1$ \\
    \midrule
    \textsc{Radar} & \cmark & \cmark & \cmark & $0.509$ & $0.262$ & $0.450$ & $0.397$ & $0.346$ & $0.393$ & $2.61$ & $0.460$ & $0.567$ & $0.627$ & $0.653$ \\
    \textsc{Backbone} & \xmark & \xmark & \xmark & $0.497$ & $0.259$ & $0.444$ & $0.396$ & $0.343$ & $0.387$ & $2.67$ & $0.402$ & $0.495$ & $0.565$ & $0.581$ \\
    \textsc{Radar}$_\text{\textit{w/o} F}$ & \xmark & \xmark & \cmark & $0.506$ & $0.260$ & $0.448$ & $0.396$ & $0.343$ & $0.391$ & $2.63$ & $0.442$ & $0.545$ & $0.624$ & $0.651$ \\
    \textsc{Radar}$_\text{\textit{w/o} SF}$ & \cmark & \xmark & \cmark & $0.508$ & $0.262$ & $0.451$ & $0.398$ & $0.346$ & $0.394$ & $2.62$ & $0.447$ & $0.543$ & $0.626$ & $0.650$ \\
    \textsc{Radar}$_\text{\textit{w/o} PF}$ & \xmark & \cmark & \cmark & $0.508$ & $0.261$ & $0.450$ & $0.396$ & $0.344$ & $0.389$ & $2.63$ & $0.456$ & $0.559$ & $0.623$ & $0.652$ \\
    \bottomrule
    \end{tabular}}
    \caption{Ablation results of \textsc{Radar} with different modules. Per-observation results of \textsc{Backbone}, \textsc{Radar}$_\text{\textit{w/o} F}$, \textsc{Radar}$_\text{\textit{w/o} SF}$, \textsc{Radar}$_\text{\textit{w/o} PF}$, and \textsc{Radar} are provided in Appendix, Table \ref{table: ablation_observation}.}
    \label{table: mimic_ablation}
\end{table*}

\begin{table*}[t]
    \centering
    \resizebox{\textwidth}{!}{
    \begin{tabular}{l|ccc|cccc|ccccccc}
    \toprule
    \multirow{2}{*}{\textbf{Model}} & \multicolumn{3}{c|}{\textbf{Modules}} & \multicolumn{4}{c|}{\textbf{Lexical Metrics}} & \multicolumn{7}{c}{\textbf{Clinical Metrics} (\textit{CheXpert: Uncertain as Negative})} \\
    & \textbf{Vision} & \textbf{Resampler} & \textbf{LLM} & \textbf{B-1} & \textbf{B-4} & \textbf{MTR} & \textbf{R-L} & \textbf{RG-F}$_1$ & \textbf{RG}$_\text{ER}$ & \textbf{CliQ}$_\text{0}$ ($\downarrow$) & \textbf{$^{14}$Ma-F}$_1$ & \textbf{$^{5}$Ma-F}$_1$ & \textbf{$^{14}$Mi-F}$_1$ & \textbf{$^{5}$Mi-F}$_1$ \\ \midrule
    \textsc{Backbone} & \cmark & \cmark & \cmark & $0.497$ & $0.259$ & $0.444$ & $0.396$ & $0.343$ & $0.387$ & $2.67$ & $0.402$ & $0.495$ & $0.565$ & $0.581$ \\
    \textsc{Backbone-V1} & \xmark & \cmark & \xmark & $0.430$ & $0.183$ & $0.359$ & $0.318$ & $0.245$ & $0.296$ & $3.15$ & $0.284$ & $0.415$ & $0.476$ & $0.508$\\
    \textsc{Backbone-V2} & \xmark & \cmark & \cmark & $0.483$ & $0.246$ & $0.428$ & $0.381$ & $0.321$ & $0.368$ & $2.78$ & $0.361$ & $0.465$ & $0.532$ & $0.550$ \\
    \bottomrule
    \end{tabular}}
    \caption{Ablation results of fine-tuning different modules of \textsc{Backbone}.}
    \label{table: blip3_ablation}
\end{table*}
\section{Experiments}
\subsection{Datasets}
We evaluate our model using three publicly available radiology report generation datasets: MIMIC-CXR\footnote{\url{https://physionet.org/content/mimic-cxr-jpg/2.0.0/}} \cite{mimic_cxr}, \textsc{CheXpert Plus}\footnote{\url{https://aimi.stanford.edu/datasets/chexpert-plus}} \cite{chexpert_plus}, and \textsc{IU X-ray}\footnote{\url{https://openi.nlm.nih.gov/}} \cite{iu_xray}:

\begin{itemize}
    \item {MIMIC-CXR} contains 377,110 chest radiographs and 227,827 reports. We use this dataset for fine-tuning, and we include only frontal images in our experiments. The number of samples in the train, validation, and test sets is 162,955, 1,286, and 2,461, respectively.
    \item {\textsc{CheXpert Plus}} comprises 223,462 unique radiology reports and chest X-ray pairs from 187,711 studies. We evaluate our model using only frontal images from the validation set, which includes 62 samples.
    \item {\textsc{IU X-ray}} is a dataset collected by Indiana University. Following \citet{maira2}, we use all frontal images for evaluation, totaling 3,199 studies.
\end{itemize}

\subsection{Evaluation Metrics}
\textbf{Lexical Metrics.} Following previous research \cite{r2gen,dcl}, BLEU-1/4 \cite{bleu}, ROUGE-L \cite{rouge}, and METEOR \cite{meteor} are adopted for evaluating the languages of generated outputs.

\noindent\textbf{Clinical Metrics.} We evaluate the factual accuracy using several metrics. Specifically, RG-F$_1$ and {RG}$_{\overline{\text{ER}}(\text{ER})}$ \cite{radgraph} evaluate the entity-level factuality and RadCliQ$_0$ \cite{radcliq}, denoted as CliQ$_0$, aligns with the preference of radiologists. For observation evaluation, $^{14}$Macro-F$_1$ ($^{14}$Ma-F$_1$) and $^{14}$Micro-F$_1$ ($^{14}$Mi-F$_1$) evaluate the macro and micro F$_1$ of 14 observations (refers to Table \ref{table: ablation_observation}), respectively. In addition, $^{5}$Macro-F$_1$ ($^{5}$Ma-F$_1$) and $^{5}$Micro-F$_1$ ($^{5}$Mi-F$_1$) measure the performance of 5 common observations (\textit{Atelectasis}, \textit{Cardiomegaly}, \textit{Consolidation}, \textit{Edema}, and \textit{Pleural Effusion)}. Two lines of CheXpert results are reported, i.e., \textit{Uncertain as Negative} and \textcolor{gray}{\textit{Uncertain as Positive}}.
\begin{table}[h]
    \centering
    \resizebox{\linewidth}{!}
    {
    \begin{tabular}{l|cc}
        \toprule
        \textbf{Hyperparameters} & \textbf{Stage I} & \textbf{Stage II} \\
        \midrule
        \multirow{3}{*}{Trainable Module} & {Vision Encoder (LoRA)} & \multirow{3}{*}{LLM (LoRA)} \\
        & Perceiver Resampler (Full) & \\
        & LLM (LoRA) & \\
        \midrule
        Training Epoch & $3$ & $2$ \\
        \midrule
        Learning Rate & \multicolumn{2}{c}{$1e-4$} \\
        Optimizer & \multicolumn{2}{c}{AdamW} \\
        LR Scheduler & \multicolumn{2}{c}{Cosine} \\
        Warmup Ratio & \multicolumn{2}{c}{$0.03$} \\
        LoRA Config & \multicolumn{2}{c}{$r=64, \alpha=128$} \\
        Batch Size & \multicolumn{2}{c}{$32$} \\
        \bottomrule
    \end{tabular}
    }
    \caption{Detailed hyperparameters for training \textsc{Radar}. LoRA is used to fine-tune both the vision encoder and the LLM, while the Perceiver Resampler is fully fine-tuned.}
    \label{table: hyperparam}
\end{table}

\subsection{Baselines}
On the MIMIC-CXR dataset, we compare our models with the state-of-the-art (SOTA) MLLMs, including {RadFM} \cite{radfm}, {XrayGPT} \cite{xraygpt}, {LLaVA-Med} \cite{llavamed}, R2GenGPT \cite{r2gengpt}, {R2-LLM} \cite{r2llm}, {RaDialog} \cite{radialog}, {CheXagent} \cite{chexagent}, {GPT-4V} \cite{gpt4v}, {LLaVA-Rad} \cite{llavarad}, {Med-PaLM} \cite{medpalm}, {MAIRA-1} \cite{maira1}, {MAIRA-2} \cite{maira2}, {MedVerse} \cite{medverse}, M4CXR \cite{m4cxr}, and {Libra} \cite{libra}. Other SOTA specialists are in the Appendix \ref{sota_specialists}. We also compare \textsc{Radar} with LLaVA-Rad and MAIRA-2 on the \textsc{IU X-ray} dataset. On the \textsc{CheXpert-Plus} dataset, we compare \textsc{Radar} with the baseline \textsc{Swin$_\text{v2}$-BERT} \cite{chexpert_plus} consisting of a Swin Transformer V2 \cite{swintransformer_v2} and a BERT decoder \cite{bert}. The \textsc{Swin$_\text{v2}$-BERT} model has three variants, each trained on a different dataset: the MIMIC-CXR dataset, the \textsc{CheXpert Plus} dataset, and a combined version of both. 

\subsection{Implementation Details}
\noindent\textbf{Training and Inference.} We implement \textsc{Radar} using BLIP-3\footnote{The model card is "Salesforce/xgen-mm-phi3-mini-instruct-interleave-r-v1.5."} \cite{blip3} as the backbone, which comprises a SigLIP \cite{siglip} vision encoder, a Perceiver Resampler, and a Phi-3-mini$_\text{3.8B}$ \cite{phi3} language model. Our implementation is based on Hugging Face’s \texttt{Transformers} library \cite{huggingface}. The expert model consists of a Swin Transformer V2\footnote{The model card is "microsoft/swinv2-large-patch4-window12to16-192to256-22kto1k-ft."} \cite{swintransformer_v2} and a  BioClinicalBERT\footnote{The model card is "emilyalsentzer/Bio\_ClinicalBERT."} \cite{alsentzer2019publiclyavailableclinicalbert}. Top-$2$ reports are selected as knowledge. The hyperparameters used for training \textsc{Radar} are provided in Table \ref{table: hyperparam}. During inference, we employ beam search with a beam width of $5$ for report generation and set the length penalty to $2.0$. As proposed by \citet{blip3}, BLIP-3 samples vision tokens using a Perceiver Resampler with learned queries and supports images of any resolution, resulting in significant performance gains across multiple tasks. In this paper, we use only the base resolution ($384 \times 384$) with $128$ learned query tokens to ensure a fair comparison with other baselines. For training, in Stage I, we fine-tune all three components (i.e., the vision encoder, the Perceiver Resampler, and the LLM) in BLIP-3 since the model is not specifically designed for medical tasks. In Stage II, we further fine-tune only the LoRA of the LLM to enhance performance.

\noindent\textbf{Data Preprocessing.} Following previous research \cite{maira1,maira2,libra}, we incorporate \textit{Indication}, \textit{History}, \textit{Comparison}, \textit{Technique}, and \textit{Prior Findings} as clinical context for the MIMIC-CXR and \textsc{CheXpert Plus} datasets, when available. Since the \textsc{IU X-ray} dataset does not include follow-up studies, we extract only \textit{Indication}, \textit{Comparison}, and \textit{Technique} as clinical context. For a better illustration, we provide the prompt template in Table \ref{table: prompt_template}.

\section{Results and Analyses}
\subsection{Quantitative Analysis}
\textbf{Comparison with MLLMs.} As shown in Table \ref{table: mllm_mimic_frontal_results}, \textsc{Radar} achieves SOTA performance compared to other MLLM baselines. In terms of lexical metrics, \textsc{Radar} outperforms the best baselines (i.e., Libra and MAIRA-2) with absolute improvements of $1.7\%$ in BLEU-4 and $1.3\%$ in ROUGE-L, while maintaining competitive performance of $0.509$ in BLEU-1 and $0.450$ in METEOR. Regarding entity-level clinical metrics, our model achieves the best performance on RG-F$_1$ and RadCliQ$_0$, attaining scores of $0.346$ and $2.61$, respectively. Additionally, \textsc{Radar} surpasses the top baselines, achieving improvements across multiple observation-level clinical metrics, with $^{14}$Macro-F$_1$ increasing to $0.460$, $^{5}$Macro-F$_1$ to $0.567$, $^{14}$Micro-F$_1$ to $0.627$, and $^{5}$Micro-F$_1$ to $0.653$, respectively. Notably, the smallest gain over the second-best model is $2.1\%$, underscoring \textsc{Radar}'s effectiveness. Furthermore, we provide an additional set of CheXpert results using the \textcolor{gray}{\textit{Uncertain as Positive}} policy and compare \textsc{Radar} with MAIRA-1. We observe that the improvements under this setting follow a similar trend to those obtained with the \textit{Uncertain as Negative} policy. These results collectively demonstrate the effectiveness of \textsc{Radar} in generating coherent and clinically accurate radiology reports.

\noindent\textbf{Comparison with SOTA Specialists.} The results of other specialists are shown in Table \ref{table: full_specialist_results}. We find that models incorporating clinical context (e.g., \textit{Indication}) as input generally achieve better performance than others. For example, the Controllable model significantly outperforms other baselines across both lexical and clinical metrics. This trend also holds for MLLMs, as shown in Table \ref{table: mllm_mimic_frontal_results}. Moreover, benefiting from the strong contextual comprehension and language generation capabilities of LLMs, \textsc{Radar} further improves linguistic quality, which requires models to integrate diverse information sources. However, we observe that the $^{14}$Macro-F$_1$ score of our model still lags behind that of the Controllable baseline ($0.497$ vs. $0.553$). This discrepancy may stem from differences in learning objectives, as this baseline treats \textit{Uncertain} cases as \textit{Positive}.

\noindent\textbf{Model Generalization.}  
Following prior research \cite{maira2}, we further evaluate \textsc{Radar} on the \textsc{CheXpert Plus} and \textsc{IU X-ray} datasets to assess its generalization capability. The results are presented in Table \ref{table: iu_xray_results} and Table \ref{table: chexpert_plus_results}. On the \textsc{IU X-ray} dataset, \textsc{Radar} outperforms MAIRA-2 in terms of CheXpert metrics, achieving a $^{14}$Macro-F$_1$ of $0.325$ and a $^{14}$Micro-F$_1$ of $0.546$. However, a performance gap remains in RG-F$_1$ and RadCliQ$_0$, which may be attributed to differences in training data, as MAIRA-2 is trained with the additional USMix dataset. Meanwhile, \textsc{Radar} demonstrates comparable performance to the baselines in terms of lexical metrics. On the \textsc{CheXpert Plus} dataset, our model significantly outperforms \textsc{Swin$_\text{v2}$-BERT} trained on the \textsc{MIMIC-CXR} dataset, across both lexical and clinical metrics. Furthermore, \textsc{Radar} surpasses the baseline that is trained on \textsc{CheXpert Plus} alone as well as the one trained on a combination of both datasets. These results demonstrate the strong generalization ability of \textsc{Radar} across different datasets. Additionally, \textsc{Radar} significantly outperforms the \textsc{Backbone}, underscoring the effectiveness of the integrated knowledge.

\begin{figure}[t]
    \centering
    \includegraphics[width=\linewidth]{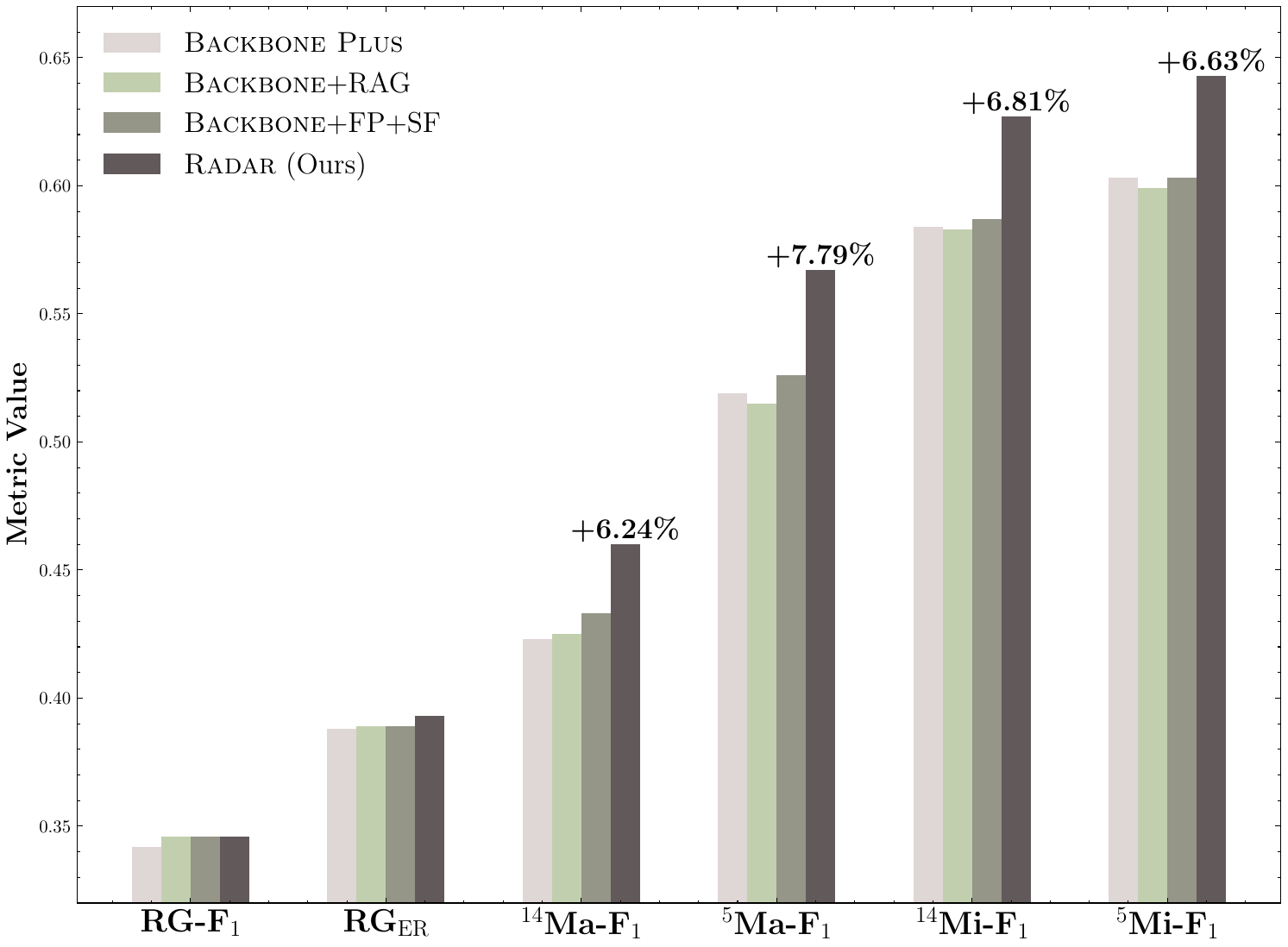}
    \caption{Comparisons among \textsc{Backbone}+RAG, \textsc{Backbone}+FP+SF, and \textsc{Radar} on six clinical metrics.}
    \label{figure: ablation_rag}
\end{figure}

\begin{figure*}[t]
    \centering
    \includegraphics[width=1.0\linewidth]{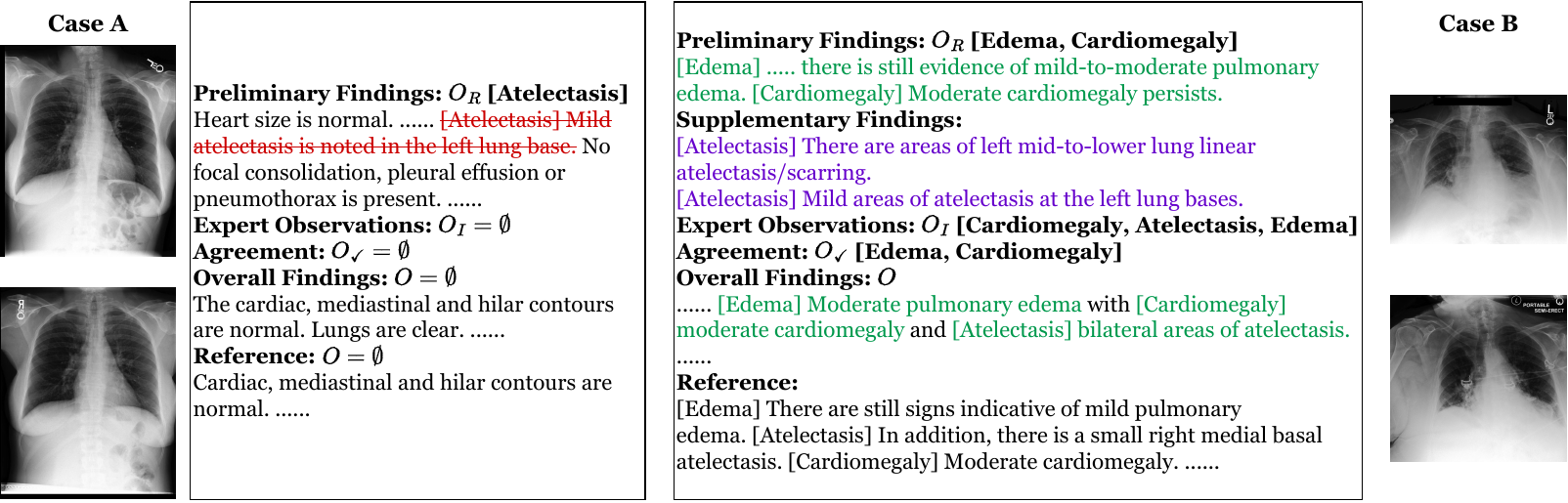}
    \caption{Two cases generated by \textsc{Radar}, where false positive observation appears in the PF of case A and false negative observation shows in the PF of case B.}
    \label{case_study}
\end{figure*}

\noindent\textbf{Analysis of PF, SF, and OI.} We analyze the impact of PF, SF, and OI on the performance of \textsc{Radar}, with results summarized in Table \ref{table: mimic_ablation}. \textsc{Radar}$_\text{\textit{w/o} F}$, which first identifies observations before report generation without incorporating knowledge, significantly improves the CheXpert metrics, particularly $^{14}$Macro-F$_1$ and $^{5}$Macro-F$_1$, as observation information captures high-level abstractions of reports and aligns closely with the objectives of these metrics. This highlights the crucial role of OI in enhancing clinical accuracy, independent of other components. When PF and SF are introduced individually with OI, introducing PF alone helps preserve the knowledge embedded in the LLM, resulting in comparable performance across both lexical and clinical metrics. In contrast, introducing SF alone substantially improves $^{14/5}$Macro-F$1$, but negatively impacts RG$_\text{ER}$ and RadCliQ$_0$. Moreover, combining both PF and SF leverages the strengths of each, leading to further improvements in the clinical metrics while maintaining comparable performance across the other metrics. We notice that \textsc{Backbone} tends to retain easily acquired knowledge (i.e., PF) and that selectively supplementing it with external information (i.e., SF) is crucial for bridging the remaining knowledge gaps. 

\noindent\textbf{Analysis of \textsc{Radar} versus RAG.} To evaluate the effectiveness of knowledge integration in \textsc{Radar}, we conduct experiments comparing our model against three baselines: (1) \textsc{Backbone Plus}, (2) \textsc{Backbone+RAG}, and (3) \textsc{Backbone}+PF+SF. The results are presented in Figure \ref{figure: ablation_rag}. Note that these baselines do not include the OI. Since \textsc{Radar} undergoes two-stage training (i.e., two additional epochs), we apply the same extended training to \textsc{Backbone}, referring to this variant as \textsc{Backbone Plus}. In addition, we introduce a standard RAG baseline (\textsc{Backbone+RAG}), which utilizes the same retrieved findings as \textsc{Radar}. Building upon this baseline, \textsc{Backbone}+PF+SF further includes PF as context. Our findings reveal that while all four models achieve comparable performance on lexical metrics (e.g., 50\%/26\% B-1/4), they differ in clinical metrics. Specifically, \textsc{Backbone+RAG} and \textsc{Backbone Plus} show similar performance, and \textsc{Backbone+FP+SF} outperforms these two baselines on CheXpert metrics and exhibits similar performance on RadGraph metrics. This demonstrates that incorporating credible knowledge can effectively enhance report generation even without OI. Moreover, \textsc{Radar} demonstrates a relative improvement of over 6\% across four key CheXpert metrics. This suggests that structured integration of internal and external knowledge contributes to its enhanced clinical accuracy.

\noindent\textbf{Analysis of Fine-tuning Different Modules in \textsc{Backbone}.} To assess the contributions of different components in the base model (i.e., BLIP-3), we conduct an ablation study on the impact of fine-tuning the vision encoder, the Resampler, and the LLM. The results are summarized in Table \ref{table: blip3_ablation}. By comparing \textsc{Backbone} and \textsc{Backbone-V2}, we find that fine-tuning the vision encoder to incorporate domain-specific knowledge is crucial for achieving high clinical accuracy, even though both configurations exhibit strong language coverage in lexical metrics. Furthermore, fine-tuning the LLM (i.e., Phi-3) results in substantial improvements in both lexical and clinical metrics, as evidenced by the comparison between V1 and V2. This highlights the importance of adapting the LLM to the clinical domain for optimal performance. Notably, \textsc{Radar} utilizes a 3.8B LLM as the decoder and outperforms many larger models (e.g., LLaVA-Med and MAIRA-1).

\subsection{Qualitative Analysis}
\textbf{Case Study.} We conduct a case study to illustrate the advantages of incorporating both internal knowledge and retrieved information, as shown in Figure \ref{case_study}. In Case A, \textsc{Radar} initially generates a report that includes the finding \textit{Atelectasis}. However, expert assessment indicates the image shows no positive findings. As a result, their intersection is $\emptyset$, and by removing this incorrect observation, \textsc{Radar} ultimately produces an accurate report. This example highlights the model’s ability to refine its predictions when guided by expert constraints, effectively eliminating unnecessary or incorrect findings. Another more complex case is presented on the right side of this Figure. Specifically, \textsc{Radar} initially identifies findings related to \textit{Edema} and \textit{Cardiomegaly}, which the expert model also notes. However, the observation of \textit{Atelectasis} is omitted from the preliminary findings. By incorporating retrieved evidence such as \textit{"... linear atelectasis ..."} and \textit{"Mild areas of atelectasis ..."}, \textsc{Radar} successfully corrects the omission and generates a complete and accurate report. This case demonstrates the model’s capability to leverage external knowledge to recover missing findings, thereby improving factual completeness.

\begin{figure}[t]
    \centering
    \includegraphics[width=\linewidth]{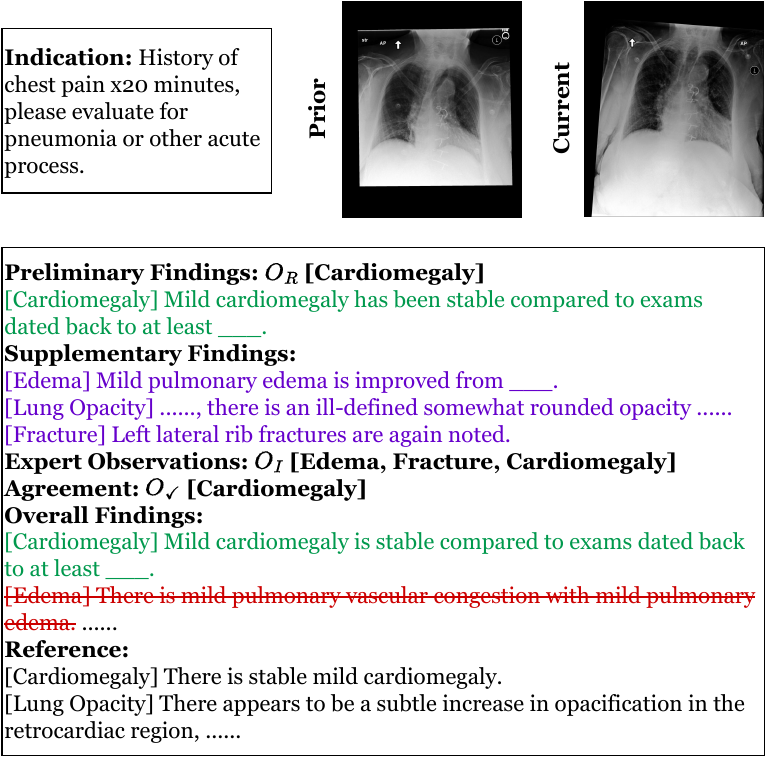}
    \caption{Error case generated by \textsc{Radar}, where \sout{\textcolor{red}{spans}} and \textcolor{green}{spans} indicate incorrect and correct observations.}
    \label{error_case}
\end{figure}

\noindent\textbf{Error Analysis.} We conduct an error analysis to gain deeper insights, as shown in Figure~\ref{error_case}. \textsc{Radar} initially generates a report containing the observation \textit{Cardiomegaly}, which is also present in the expert model’s output. In this case, the observation reflects credible knowledge possessed by the LLM and should be preserved. Subsequently, \textsc{Radar} produces a false positive finding, \textit{Edema}, which aligns with the retrieved supplementary findings. This error may result from the model’s overreliance on external knowledge. Moreover, since \textit{Edema} is clinically associated with \textit{Cardiomegaly}, it is possible that \textsc{Radar} has learned only superficial correlations between them. To address these issues, potential solutions include refining the expert model and expanding the training dataset.
\section{Related Works}
Radiology report generation \cite{coatt,hrgr} is a valuable yet challenging task. Numerous research efforts have been dedicated to improving clinical accuracy, employing diverse approaches such as memory-based neural models \cite{r2gen,r2gencmn}, planning-based methods \cite{coplan,organ}, and reinforcement learning-optimized techniques \cite{clinically_coherent,fact_ent,cmm-rl}. Additionally, several studies \cite{ramesh2022improving,bannur2023learning,recap,serra-etal-2023-controllable,icon} have addressed the issue of hallucination, particularly in the absence of prior studies. Given the critical role of domain knowledge in this field, researchers have leveraged knowledge graphs to enhance report generation \cite{mia, dcl, kiut, yan-etal-2023-style}.  

With the emergence of MLLMs \cite{blip2,llava}, which demonstrate exceptional capabilities in image understanding and captioning, many studies \cite{medpalm,radfm,xraygpt,llavamed} have explored their application in the medical domain. \citet{chexagent} introduced a foundation model for chest X-ray interpretation, while \citet{llavarad} developed a lightweight MLLM tailored for radiology. \citet{m4cxr} investigated the multi-task potential of LLMs, and \citet{libra} incorporated temporal information to enhance chest X-ray analysis.
\section{Conclusion}  
In this paper, we introduce \textsc{Radar}, a novel approach designed to enhance radiology report generation by leveraging both the internal knowledge of an LLM and externally retrieved information. Our model first generates a report and subsequently classifies the image based on observations, with their shared components regarded as internal knowledge. It then retrieves supplementary information to further refine and complement this knowledge. Extensive experiments on three public datasets demonstrate that \textsc{Radar} achieves SOTA performance in both language quality and clinical accuracy, highlighting the effectiveness of integrating internal and external knowledge for more accurate and coherent radiology report generation.
\section*{Limitations}
Our experiments are conducted using a single backbone architecture. While this choice provides a controlled evaluation, the performance of alternative architectures remains unexplored. Future work should investigate whether different model architectures can achieve comparable or better results. In addition, our study focuses exclusively on a single imaging modality (e.g., Chest X-ray). The model's effectiveness in other imaging modalities, such as CT scans or MRI, has not been evaluated. Extending our approach to multiple imaging modalities would be an important direction for future research to enhance its clinical utility and generalizability.
\section*{Ethical Considerations}
This study utilizes the MIMIC-CXR \cite{mimic_cxr}, \textsc{IU X-ray} \cite{iu_xray}, and \textsc{CheXpert Plus} \cite{chexpert_plus} datasets, all of which are publicly available and have been automatically de-identified to mitigate privacy risks. Our primary objective is to improve the clinical accuracy of reports generated by LLMs in medical imaging. However, despite our efforts, the generated reports may contain inaccuracies or omissions. Therefore, these outputs should not be used as a substitute for expert medical judgment. We strongly advocate for thorough validation by qualified radiologists or healthcare professionals before any clinical or diagnostic application.

\bibliography{custom}

\appendix
\section{Appendix}
\subsection{Full List of Specialists}\label{sota_specialists}
In addition to specialist baselines in Table \ref{table: mllm_mimic_frontal_results}, the following baselines are included: \textsc{R2Gen} \cite{r2gen}, \textsc{R2GenCMN} \cite{r2gencmn}, $\mathcal{M}^2$\textsc{Tr} \cite{m2tr}, \textsc{KnowMat} \cite{mia}, \textsc{CMM-RL} \cite{cmm-rl}, CMCA \cite{cmca}, KiUT \cite{kiut}, DCL \cite{dcl}, METrans \cite{metrans}, RGRG \cite{rgrg}, \textsc{Recap} \cite{recap}, Controllable \cite{serra-etal-2023-controllable}, PromptMRG \cite{promptingmrg}, and \textsc{ICon} \cite{icon}.

\begin{table}[h]
\centering
    \resizebox{\linewidth}{!}
    {
    \begin{tabular}{l|ccc}
    \toprule
    \textbf{Observation} & \textbf{P} & \textbf{R} & \textbf{F${_1}$} \\
    \midrule
    \textit{Atelectasis} & $0.518$ & $0.645$ & $0.574$ \\
    \textit{Cardiomegaly} & $0.656$ & $0.783$ & $0.713$ \\
    \textit{Consolidation} & $0.370$ & $0.174$ & $0.237$ \\
    \textit{Edema} & $0.518$ & $0.610$ & $0.560$ \\
    \textit{Pleural Effusion} & $0.695$ & $0.800$ & $0.744$ \\
    \midrule
    \rowcolor{lightgray!50}
    $^5$Macro Average & $0.551$ &$0.602$ & $0.567$ \\
    \rowcolor{lightgray!50}
    $^5$Micro Average & $0.607$ & $0.707$ & $0.653$ \\
    \midrule
    \textit{Enlarged Card.} & $0.277$ & $0.204$ & $0.235$ \\
    \textit{Lung Opacity} & $0.644$ & $0.496$ & $0.561$ \\
    \textit{Lung Lesion} & $0.492$ & $0.207$ & $0.291$ \\
    \textit{Pneumonia} & $0.283$ & $0.232$ & $0.255$ \\
    \textit{Pneumothorax} & $0.407$ & $0.636$ & $0.496$ \\
    \textit{Pleural Other} & $0.333$ & $0.173$ & $0.228$ \\
    \textit{Fracture} & $0.421$ & $0.244$ & $0.309$ \\
    \textit{Support Devices} & $0.823$ & $0.866$ & $0.844$ \\
    \textit{No Finding} & $0.302$ & $0.569$ & $0.395$ \\
    \midrule
    \rowcolor{lightgray!50}
    $^{14}$Macro Average & $0.481$ & $0.474$ & $0.460$ \\
    \rowcolor{lightgray!50}
    $^{14}$Micro Average & $0.614$ & $0.640$ & $0.627$ \\
    \bottomrule
    \end{tabular}
    }
    \caption{Experimental results of \textsc{Radar} for each observation on the \textsc{MIMIC-CXR} dataset.}
    \label{table: ablation_observation}
\end{table}

\begin{table*}[t]
    \centering
    \begin{tabular}{l|cccccc|ccc}
    \toprule
    \multicolumn{10}{c}{\textbf{Dataset: MIMIC-CXR} \textit{(Compared with SOTA Specialists)}} \\
    \midrule
    \multirow{2}{*}{\textbf{Model}} & \multicolumn{6}{c|}{\textbf{Lexical Metrics}} & \multicolumn{3}{c}{\textbf{CE ($^{14}$Macro) Metrics}} \\
    & \textbf{B-1} & \textbf{B-2} & \textbf{B-3} & \textbf{B-4} & \textbf{MTR} & \textbf{R-L} & \textbf{P} & \textbf{R} & \textbf{F}$_1$ \\ 
    \midrule
    \textsc{R2Gen} & $0.353$ & $0.218$ & $0.145$ & $0.103$ & $0.142$ & $0.270$ & $0.333$ & $0.273$ & $0.276$ \\
    \textsc{R2GenCMN} & $0.353$ & $0.218$ & $0.148$ & $0.106$ & $0.142$ & $0.278$ & $0.344$ & $0.275$ & $0.278$ \\
    $\mathcal{M}^2$\textsc{Tr} & $0.378$ & $0.232$ & $0.154$ & $0.107$ & $0.145$ & {$0.272$} & $0.240$ & {$0.428$} & $0.308$ \\
    \textsc{KnowMat} & $0.363$ & $0.228$ & $0.156$ & $0.115$ & $-$ & $0.284$ & {$0.458$} & $0.348$ & $0.371$ \\
    \textsc{CMM-RL} & {$0.381$} & $0.232$ & $0.155$ & $0.109$ & {$0.151$} & {$0.287$} & $0.342$ & $0.294$ & $0.292$ \\
    \textsc{CMCA} & $0.360$ & $0.227$ & {$0.156$} & $0.117$ & $0.148$ & {$0.287$} & {${0.444}$} & $0.297$ & $0.356$ \\
    KiUT & {$0.393$} & $0.243$ & $0.159$ & $0.113$ & {$0.160$} & $0.285$ & $0.371$ & $0.318$ & $0.321$ \\
    DCL & $-$ & $-$ & $-$ & $0.109$ & $0.150$ & $0.284$ & $0.471$ & $0.352$ & {$0.373$} \\
    METrans & $0.386$ & {$0.250$} & {$0.169$} & {$0.124$} & $0.152$ & {$0.291$} & $0.364$ & $0.309$ & $0.311$ \\
    RGRG & $0.373$ & $0.249$ & $0.175$ & ${0.126}$ & $0.168$ & $0.264$ & $0.380$ & $0.319$ & $0.305$ \\
    \textsc{ORGan} & {$0.407$} & {$0.256$} & {$0.172$} & $0.123$ & {$0.162$} & ${0.293}$ & \textcolor{gray}{$0.416$} & \textcolor{gray}{$0.418$} & \textcolor{gray}{$0.385$} \\
    \textsc{Recap} & {$0.429$} & ${0.267}$ & ${0.177}$ & ${0.125}$ & ${0.168}$ & $0.288$ & \textcolor{gray}{$0.389$} & \textcolor{gray}{$0.443$} & \textcolor{gray}{$0.393$} \\
    Controllable & $\underline{0.486}$ & $\underline{0.366}$ & $\underline{0.295}$ & $\underline{0.246}$ & $\underline{0.216}$ & $\bm{0.423}$ & \textcolor{gray}{$\bm{0.597}$} & \textcolor{gray}{$\bm{0.516}$} & \textcolor{gray}{$\bm{0.553}$}\\
    PromptMRG & $0.398$ & $-$ & $-$ & $0.112$ & $0.157$ & $0.268$ & $0.396$ & $0.393$ & $0.381$ \\
    \textsc{ICon} & $0.429$ & $0.266$ & $0.178$ & $0.126$ & $0.170$ & $0.287$ & \textcolor{gray}{$0.445$} & \textcolor{gray}{$0.505$} & \textcolor{gray}{$0.464$} \\
    \midrule
    \textsc{Radar} (Ours) & $\bm{0.509}$ & $\bm{0.390}$ & $\bm{0.315}$ & $\bm{0.262}$ & $\bm{0.450}$ & $\underline{0.397}$ & \makecell{$\bm{0.481}$\\\textcolor{gray}{$\underline{0.523}$}} & \makecell{$\bm{0.474}$\\\textcolor{gray}{$\underline{0.500}$}} & \makecell{$\bm{0.460}$\\\textcolor{gray}{$\underline{0.497}$}} \\
    \bottomrule
    \end{tabular}
    \caption{Experimental results of our model and SoTA specialists on the \textsc{MIMIC-CXR} dataset. \textcolor{gray}{Results denotes \textit{Uncertain} as \textit{Positive}.}}
    \label{table: full_specialist_results}
\end{table*}

\begin{table*}[t]
\centering
    \resizebox{\textwidth}{!}
    {
    \begin{tabular}{l|l}
    \toprule
    \textbf{Role} & \textbf{Prompt} \\
    \midrule
    \multirow{3}{*}{\textsc{System}} & \texttt{<|system|>} \\
    & \multicolumn{1}{p{13cm}}{You are an assistant in radiology, responsible for analyzing medical imaging studies and generating detailed, structured, and accurate radiology reports.} \\
    & \texttt{<|end|>} \\
    \midrule
    \multirow{10}{*}{\textsc{User}} & \texttt{<|user|>} \\
    & {\color{gray}\texttt{<prior image>} (\textit{If prior available})} \\
    & \texttt{<current image>} \\
    & \textit{Indication}: ......\\
    & \textit{History}: ......\\
    & \textit{Comparison}: ......\\
    & \textit{Technique}: ......\\
    & {\color{gray}\textit{Prior Findings}: ...... (\textit{If prior available})} \\
    & {\color{gray}\textit{Preliminary Findings}: ...... (\textit{If available})} \\
    & {\color{gray}\textit{Supplementary Findings}: ...... (\textit{If available})} \\
    & \multicolumn{1}{p{13cm}}{Generate a comprehensive and detailed description of findings based on this chest X-ray image. {\color{gray} Include a thorough comparison with a prior chest X-ray, emphasizing any significant changes, progression, or improvement. (\textit{If prior available})} {\color{gray} Before this, systematically identify all observations.}}\\
    & \texttt{<|end|>} \\
    \midrule
    \multirow{6}{*}{\textsc{Assistant}} & \texttt{<|assisitant|>} \\
    & \textit{Identified Observations}: \\
    & ...... \\
    & \textit{Overall Findings}: {\color{gray} \textit{(e.g., the target)}}\\
    & ...... \\
    & \texttt{<|end|>} \\
    \bottomrule
    \end{tabular}
    }
    \caption{The prompt template for \textsc{Radar} and its variants, consisting of three roles: System, User, and Assistant.}
    \label{table: prompt_template}
\end{table*}

\end{document}